\documentclass[11pt, a4paper, logo, onecolumn,copyright]{tmlrmel}

\usepackage[authoryear, sort&compress, round]{natbib}
\title{Distributional Statistics Restore Training Data Auditability in One-step Distilled Diffusion Models}

\usepackage[utf8]{inputenc} 
\usepackage[T1]{fontenc}    
\usepackage{hyperref}       
\usepackage{url}            
\usepackage{booktabs}       
\usepackage{nicefrac}       
\usepackage{microtype}      
\usepackage{amsmath}
\usepackage{graphicx}
\usepackage{multicol}
\usepackage{siunitx}
\usepackage{array}
\usepackage[nameinlink]{cleveref}
\usepackage{bbm}
\usepackage{multirow}
\usepackage{subfig}
\usepackage{soul}
\usepackage{floatrow}
\usepackage{float}
\usepackage{wrapfig}
\usepackage{blindtext}
\usepackage{tablefootnote}
\usepackage{amsfonts}
\usepackage[flushleft]{threeparttable}
\usepackage{colortbl}
\usepackage{bbding}
\usepackage{lipsum}
\usepackage[toc,page,header]{appendix}
\usepackage{algorithmic}
\usepackage{algorithm}
\theoremstyle{plain}

\theoremstyle{definition}

\theoremstyle{remark}

\usepackage{thm-restate}

\usepackage{xcolor}
\usepackage{colortbl}
\definecolor{lg}{gray}{0.9}
\definecolor{dg}{RGB}{0,150,0}
\definecolor{dr}{RGB}{139,0,0}

\usepackage{xspace}
\usepackage{floatrow}

\newcommand{\draftonly}[1]{#1}
\newcommand{\eat}[1]{}
\renewcommand{\draftonly}[1]{}
\definecolor{darkgreen}{RGB}{0, 102, 0}

\crefformat{section}{\S#2#1#3}

\usepackage{amsmath,amsfonts,bm}

\def\eqref#1{equation~\ref{#1}}

\def\1{\bm{1}}

\def\vx{{\bm{x}}}

\DeclareMathAlphabet{\mathsfit}{\encodingdefault}{\sfdefault}{m}{sl}
\SetMathAlphabet{\mathsfit}{bold}{\encodingdefault}{\sfdefault}{bx}{n}

\def\gB{{\mathcal{B}}}

\def\gD{{\mathcal{D}}}

\def\gL{{\mathcal{L}}}

\def\gS{{\mathcal{S}}}

\def\gX{{\mathcal{X}}}

\def\gZ{{\mathcal{Z}}}

\def\sH{{\mathbb{H}}}
\def\sI{{\mathbb{I}}}

\def\sP{{\mathbb{P}}}
\def\sQ{{\mathbb{Q}}}
\def\sR{{\mathbb{R}}}

\author[$\diamondsuit$]{Muxing Li$^{*}$}
\author[$\diamondsuit$]{Zesheng Ye$^{*}$}
\author[$\heartsuit$]{Yixuan Li}
\author[$\spadesuit$]{Andy Song}
\author[$\clubsuit$]{Guangquan Zhang}
\author[$\diamondsuit$]{Feng Liu}

\affil[ \hspace{-0.2em}]{$\diamondsuit$ University of Melbourne
}
\affil[ \hspace{-0.2em}]{$\heartsuit$ University of Wisconsin-Madison}
\affil[ \hspace{-0.2em}]{$\spadesuit$ Royal Melbourne Institute of Technology University}
\affil[ \hspace{-0.2em}]{$\clubsuit$ University of Technology Sydney}

\correspondingauthors{fengliu.ml@gmail.com}
\equalcontributions
\coderepo{github.com/MuxingLiUnimelb/d-utdd/}

\begin{abstract}
The proliferation of diffusion models trained on web-scale, provenance-uncertain image collections has made it essential, yet technically unresolved, to determine whether a model has learned from specific copyrighted data without authorization.
Current methods primarily rely on the {\em memorization effect}, whereby models reconstruct their training images better than unseen ones, to detect unauthorized training data on a per-instance basis.
This effect, however, vanishes under {\em distillation}, the now-dominant deployment pipeline that compresses compute-intensive {\em teacher diffusion models} into efficient {\em student one-step generators} mimicking the teacher's output for real-time user access.
As the students train exclusively on teacher-generated outputs and never directly see the teacher's original training data, they carry no per-instance memorization of that upstream data, creating a {\em model laundering} loophole that severs the auditable link between a deployed model and its upstream training data.
We nonetheless reveal that a distributional {\em memory chain} survives under distillation: the student's output distribution remains closer to the teacher's training distribution than to any non-training reference, even if no single training instance is memorized.
Exploiting this chain, we develop a {\em distributional unauthorized training data detector}, grounded in kernel-based distribution discrepancy, that determines if a candidate dataset of unknown composition is statistically aligned with the student-generated distribution more than held-out non-training datasets, thus tracing provenance back to the teacher's training data.
Evaluation across benchmarks and distillation setups confirms reliable detection even when unauthorized data forms a minority of the candidate set, establishing distribution-level auditing as a countermeasure to model laundering and a paradigm for accountable generative AI ecosystems.
\end{abstract}
\keywords{data-copyright issue, unauthorized data usage detection, model laundering}

\begin{document}

\maketitle

\section{Introduction}
Generative models are becoming part of the infrastructure that produces and transforms scientific~\citep{watson2023novo}, commercial~\citep{rombach2022high, yang2023diffusion}, and cultural media through large-scale image-generation services such as Midjourney~\citep{zierock2023leveraging} and DALL-E 2~\citep{borji2022generated}.
As deployment scales, the central question has shifted from what these models can generate to {\em what data shaped them}~\citep{carlini2021extracting}, and whether those data were used with appropriate consent and licences~\citep{sag2023copyright, jiang2023ai, shan2023glaze}.

\begin{figure}[H]
\centering
\includegraphics[width=1\linewidth]{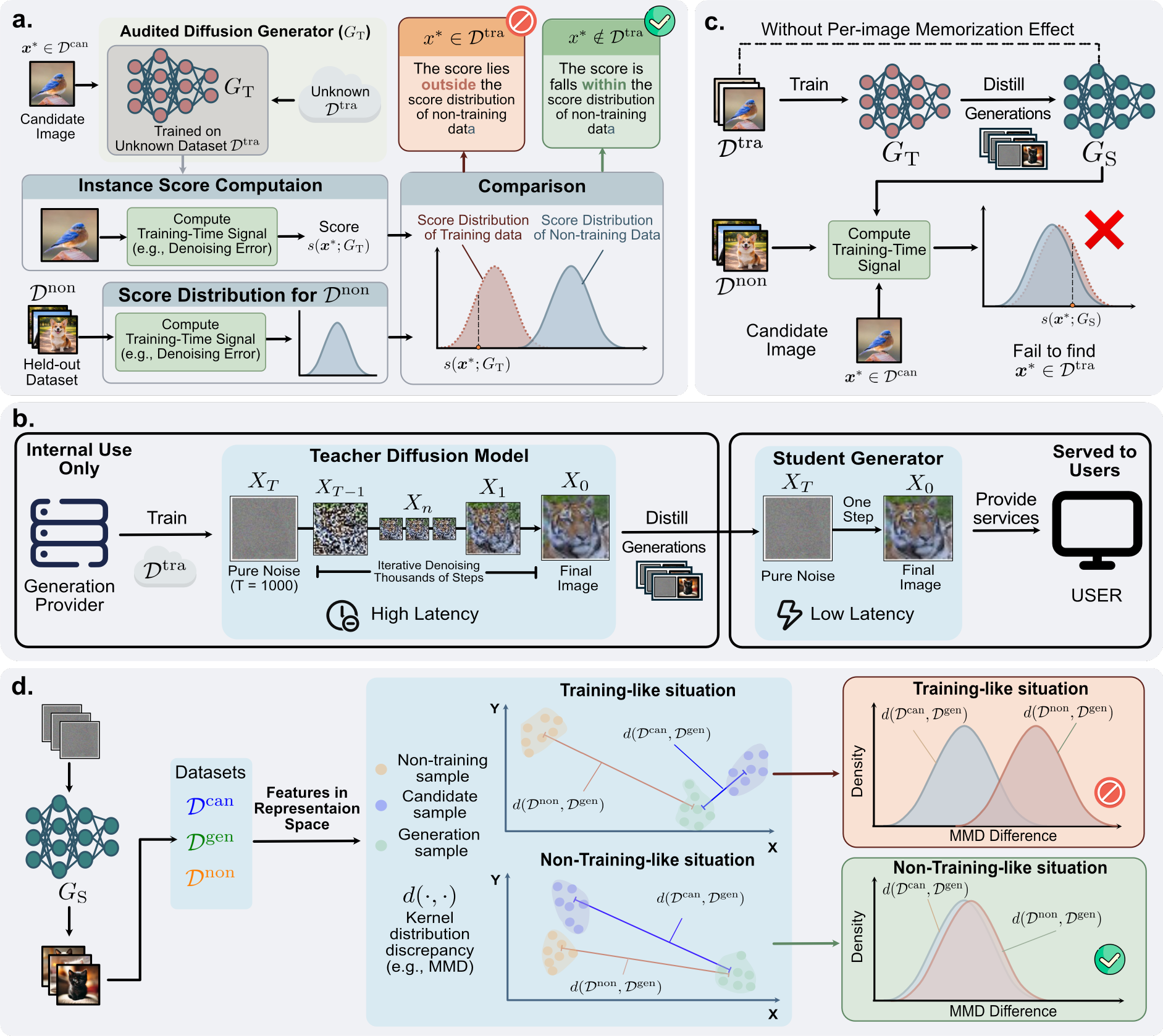}
\caption{\footnotesize \textbf{Detecting unauthorized training data under diffusion model distillation.} 
    \textbf{a}, Instance-level \textit{unauthorized training data detectors} (UTDDs). To determine if a candidate image $\vx^{*}$ was used to train a model, UTDDs compute a training-time signal $s(\vx^{*};G_{\rm T})$, directly on the audited diffusion generator $G_{\rm T}$. This score is then compared against the score distribution of a held-out non-training dataset $\mathcal{D}^{\rm non}$ to determine if $\vx^{*}$ was memorized during training.
    \textbf{b}, Diffusion Model Distillation and Deployment. To achieve low latency for users, high-latency teacher models requiring thousands of iterative denoising steps are distilled into low-latency one-step student generators. The student model is packaged and served to users, while the teacher model and its upstream training data $\mathcal{D}^{\rm tra}$ remain strictly for internal use.
    \textbf{c}, The Auditability Gap. When instance-level UTDD is applied to the deployed student model $G_{\rm S}$, it fails to identify $\vx^{*} \in \mathcal{D}^{\rm tra}$. Because $G_{\rm S}$ is trained on teacher generations rather than $\mathcal{D}^{\rm tra}$, $G_{\rm S}$ does not preserve the per-image memorization effect for $\mathcal{D}^{\rm tra}$ that instance-level UTDD relies upon. The student model effectively washes out the audit signal, creating a "model laundering" loophole.
    \textbf{d}, \textit{Distributional unauthorized training data detection} (D-UTDD). The proposed D-UTDD leverages a distribution-level memory chain. Specifically, samples from the datasets are first mapped into a shared representation space to extract high-level semantic features, avoiding the noise of raw pixel-level comparisons. It then calculates the kernel distribution discrepancy, such as \textit{Maximum Mean Discrepancy} (MMD), between student generations $\mathcal{D}^{\rm gen}$ and both the candidate set $\mathcal{D}^{\rm can}$ and the non-training set $\mathcal{D}^{\rm non}$. In a training-like setting where the candidate set was used, the candidate set exhibits a substantially smaller distributional distance to the model's generations than the non-training set.}
    \label{fig:break_chain}
\end{figure}

This provenance problem is particularly critical for diffusion models~(DMs)~\citep{ho2020denoising, karras2022elucidating, song2023consistency}, which are trained on web-scale image corpora whose provenance is uncertain and can be difficult to verify~\citep{somepalli2023diffusion, carlini2023extracting, wang2025comprehensive,li2023privacy}, necessitating audits that answer an operationally meaningful question from rights holders to image-generation providers:
\begin{center}
    {\em Was my private data used to train a deployed image generator without consent}?
\end{center}
Formally, let $G_{\rm T}$ denote an audited diffusion generator trained on an unknown dataset $\gD^{\rm tra}$ by minimizing a diffusion training objective (e.g., denoising error~\citep{ho2020denoising} or score-matching loss~\citep{song2020score}).
A rights holder (the auditor) possesses a private candidate dataset $\gD^{\rm can}$ and seeks evidence of unauthorized use---equivalently, whether $\gD^{\rm can} \cap \gD^{\rm tra} \neq \emptyset$.
To date, most {\em unauthorized training data detectors}~(UTDDs) cast this audit as a per-instance task: given a candidate image $\vx^* \in \gD^{\rm can}$, they compute a score $s(\vx^*; G_{\rm T})$ from training-time signals, such as denoising error~\citep{duan2023diffusion} or reconstruction instability~\citep{li2024towards}, then decide whether $\vx^*$ was part of $\gD^{\rm tra}$ by comparing $s(\vx^*; G_{\rm T})$ to the score distribution induced by a held-out non-training set $\gD^{\rm non}$ 
(Fig.~\ref{fig:break_chain}\textbf{a}).

The premise shared by existing UTDDs is that direct optimization on $\gD^{\rm tra}$ leaves a per-instance {\em memorization effect}---training image tend to yield systematically lower scores than held-out data~\citep{yeom2018privacy, duan2023diffusion}---making such auditing effective when the audited model has been trained directly on the contested candidate images, the setting assumed by instance‑level UTDDs~\citep{carlini2023extracting, hu2023loss, duan2023diffusion}. 

However, this setting does not account for how DMs are deployed in practice.
High-fidelity inference with DMs is compute-intensive because generation involves {\em thousands of} iterative denoising steps per image, making them costly to serve at scale.
To reduce latency and cost, practitioners increasingly distill multi-step ``teacher'' DMs into {\em one-step} ``student'' generators~\citep{kim2023consistency, luo2024diff}: the teacher is trained once, and its behavior is transferred into a student generator that produces images drastically faster while preserving much of the teacher's output quality (Fig.~\ref{fig:break_chain}\textbf{b}).

Operationally, the student is trained to mimic the teacher’s denoising trajectory~\citep{salimans2022progressive, song2023consistency}, and, in many one-step distillation pipelines, learns {\em exclusively} on teacher-generated samples, rather than the teacher's original training set~\citep{yin2024one, geng2024one}.
Distillation therefore induces a deployment strategy in which the student is the model that is packaged, served, and exposed to user access and external scrutiny, whereas the teacher remains an internal artifact used primarily for student training and model iteration~\citep{luo2023latent,li2023snapfusion,sauer2024adversarial}.

This efficiency-driven deployment transformation has an unintended consequence for compliance verification.
Rights holders who need audits interact with the deployed student interface, yet the question ``did the model memorize this image?'' they must answer concerns upstream training data.
As the student is optimized to match the teacher's output distribution, it never directly fits the teacher's training images~\citep{salimans2022progressive, yin2024one, luo2024diff} and does not develop the same per‑image memorization effect that current UTDDs rely upon.
Empirically, when only the student is accessible, state‑of‑the‑art UTDDs that succeed on the teacher collapse to near‑random performance on the distilled model and thus seem to exonerate the provider, even if the teacher was trained on unauthorized data and the developed system ultimately derives its capabilities from copyrighted data.
Since the audit interface and the provenance question are separated, the evidentiary value of UTDDs on per-instance audit appears to have been washed out, creating a {\em model laundering} loophole that risks obfuscating provenance during model life-cycle transformations in realistic scenarios, such as training on high-quality social-media imagery and then releasing only a distilled one-step generator (Fig.~\ref{fig:break_chain}\textbf{c}).
Without reliable audit signals that survive deployment transforms, the upstream training data provenance question becomes operationally unverifiable.

Still, we show that distillation does not eliminate training-data influence on student models but changes its granularity.
Across teacher-student pairs, sets of student generations $\gD^{\rm gen}$ are consistently {\em closer in distribution} to the teacher model's training set $\gD^{\rm tra}$ {\em than} to a non-training set $\gD^{\rm non}$ over repeated subset draws, as quantified by a distribution discrepancy metric called {\em maximum mean discrepancy}~(MMD)~\citep{gretton2012kernel}.
This reveals a {\em distribution-level memory chain} that links student models' behavior back to the teacher's training data and survives the distillation-based deployment transform, even when student models lack memory of such data at the instance level.

This observation motivates reframing unauthorized‑data audits from per‑image decisions to statistical evidence aggregated over datasets, namely {\em distributional unauthorized training data detection}~(D-UTDD) in a deployment setting where the auditor can access only the deployed student model $G_{\rm S}$ while the teacher $G_{\rm T}$ and its training set $\gD^{\rm tra}$ are inaccessible (Fig.~\ref{fig:break_chain}\textbf{d}).
Given a rights-holder's candidate set $\gD^{\rm can}$ (suspected to overlap with $\gD^{\rm tra}$) and a held-out reference set $\gD^{\rm non}$ known to be disjoint from $\gD^{\rm tra}$, D-UTDD draws a student-generation set $\gD^{\rm gen} \sim G_{\rm S}$ and performs relative distributional comparison: if $\gD^{\rm can}$ contains training-influenced data, the {\em distributional memory chain} discovered above implies $d(\gD^{\rm can}, \gD^{\rm gen}) < d(\gD^{\rm non}, \gD^{\rm gen})$ for a kernel distribution discrepancy $d(\cdot, \cdot)$ (e.g., MMD), because the student's output inherit the teacher's distributional bias toward $\gD^{\rm tra}$.
We instantiate $d(\cdot, \cdot)$ with a data-adaptive deep-kernel MMD~\citep{liu2020learning} computed in a representation space derived from $G_{\rm S}$, where the kernel is optimized to maximize the separation between training-like and non-training sets at the distributional level, sharpening the discriminative power of D-UTDD when unauthorized use of $\gD^{\rm can}$ indeed occurred.

Together, our results (i) show that state-of-the-art instance-level UTDDs struggle to reveal unauthorized data use in teacher DMs through their one-step student models that preserve no per-instance memorization effect, creating an auditability gap under student-only access; (ii) uncover a distribution-level memory chain that survives the teacher-to-student deployment transform, linking student outputs back to the teacher's training distribution even when instance-level signals vanish; and (iii) translate this chain into D-UTDD, enabling reliable dataset-level upstream data provenance audits across multiple teacher–student pipelines and image benchmarks, including settings where disputed data constitute a minority of the candidate dataset.
These results reposition UTDD from pinpointing memorized instances to auditing distributional provenance traces in modern generative model life‑cycles.

\section{Results}\label{sec:result}
\paragraph{Auditing provenance under model distillation.}
We study post-deployment training-data provenance audit for DM-based image generators.
In the pipeline we consider, an original DM (i.e., teacher model $G_{\rm T}$) is distilled into a one-step generator  (i.e., student model $G_{\rm S}$) trained {\em exclusively} on teacher-generated images and deployed as the user-facing model.
The auditor holds a candidate set $\gD^{\rm can}$ whose provenance is in question and a reference set $\gD^{\rm non}$ known to be absent from teacher's training set $\gD^{\rm tra}$. 
The task is to determine whether $\gD^{\rm can}$ was used to train the upstream $G_{\rm T}$, given only access to the deployed $G_{\rm S}$.

\paragraph{One-step distillation severs instance-level provenance cues.}
Recall that existing UTDDs such as SecMI~\citep{duan2023diffusion}, ReDiffuse~\citep{li2024towards} and GSA~\citep{pang2023white} rely on memorization of each instance in $\gD^{\rm tra}$ and the resulting statistical gap of score distributions between training and non-training instances to detect provenance through the deployed model $G_{\rm S}$.
However, distillation optimizes $G_{\rm S}$ to match the teacher's output distribution rather than to minimize loss on specific instances of $\gD^{\rm tra}$.
We first examined whether such detection signals remain reliable under this setup.

\begin{figure}[H]
    \centering
    \includegraphics[width=1\linewidth]{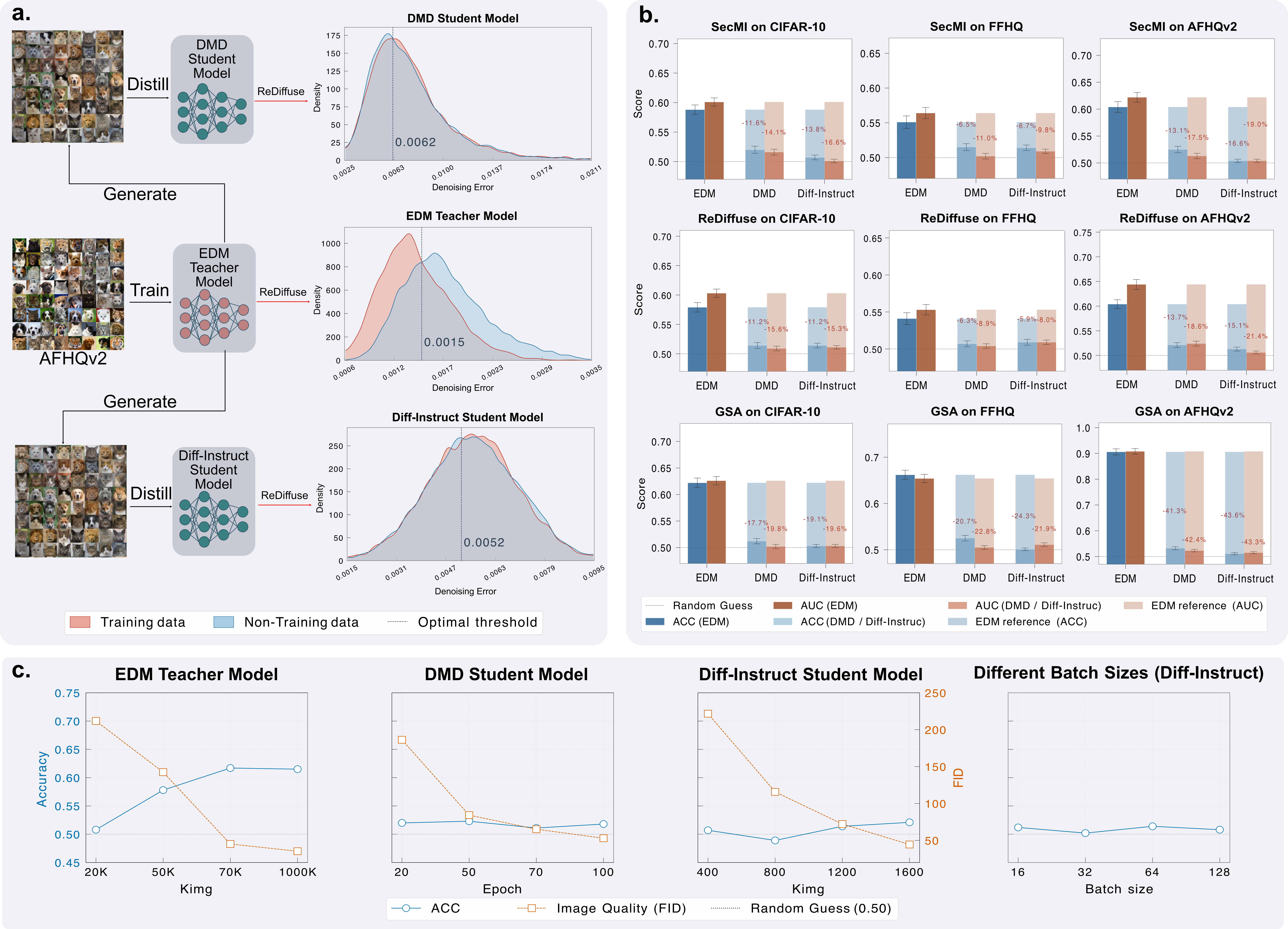}
    \caption{\textbf{The failure of instance-level UTDDs on distilled student models.} 
\textbf{a}, Distribution Difference Collapse. When evaluating training versus non-training data, instance-level UTDDs like ReDiffuse~\citep{li2024towards} compute average reconstruction error as a training-time signal. This signal forms distinctly separable distributions on the audited EDM teacher model trained on the AFHQv2 dataset~\citep{choi2020stargan}. However, after the distillation process, these score distributions overlap almost entirely on the deployed DMD and Diff-Instruct student models.
\textbf{b}, Performance Degradation Across Methods and Datasets. This auditability gap is not specific to one detector or dataset. Across multiple benchmark datasets (CIFAR-10~\citep{krizhevsky2010cifar}, FFHQ~\citep{karras2019style}, and AFHQv2) and state-of-the-art UTDDs (SecMI~\citep{duan2023diffusion}, ReDiffuse and GSA~\citep{pang2023white}), detection metrics such as Accuracy (ACC) and Area Under the Curve (AUC) perform well above chance on the EDM teacher. In contrast, performance consistently collapses to near random-guess levels for the distilled DMD and Diff-Instruct student models.
\textbf{c}, Persistence Across Training Configurations. The failure of instance-level UTDDs is mechanistic and persists regardless of the training setup. While the EDM teacher model achieves higher UTDD accuracy as training progresses over Kimg (kilo-images, measuring thousands of real images processed during training), UTDD accuracy on the student models remains near the random-guess baseline, regardless of the distillation stage or batch size.
}
    \label{fig:challenge}
\end{figure}

We evaluated two state-of-the-art distillation pipelines, DMD~\citep{yin2024one} and Diff-Instruct~\citep{luo2024diff}, derived from teacher DM architectures EDM~\citep{karras2022elucidating}.
On teacher model, ReDiffuse yields separable score distributions for training versus non-training images; on the corresponding students, the distributions overlap almost entirely {(Fig.~\ref{fig:challenge}\textbf{a})}.
Fig.~\ref{fig:challenge}\textbf{b} confirms that this collapse is not detector- nor data-specific: across benchmark datasets CIFAR-10~\citep{krizhevsky2010cifar}, FFHQ~\citep{karras2019style}, and AFHQv2~\citep{choi2020stargan}, three state-of-the-art UTDDs~\citep{duan2023diffusion, li2024towards, pang2023white} perform above chance on the teacher yet fall to near-random performance on both DMD and Diff‑Instruct.

To further examine whether this failure is tied to particular training configurations, we evaluate ReDiffuse under multiple settings, including (i) teacher models trained on AFHQv2 at different training stages, (ii) student models obtained from different stages of distillation, and (iii) student models distilled with different batch sizes (Fig.~\ref{fig:challenge}\textbf{c}).
As teacher training progresses, the model gradually achieves better generation quality and develops stronger memorization of its training data; correspondingly, ReDiffuse attains higher accuracy in identifying training instances.
In contrast, for the student models, ReDiffuse fails at every stage: regardless of the distillation stage, it consistently predicts the presence of the teacher’s training data.
Moreover, when evaluating Diff-Instruct models distilled with different batch sizes, ReDiffuse similarly exhibits consistently poor performance.

\begin{figure}[H]
    \centering
    \includegraphics[width=1\linewidth]{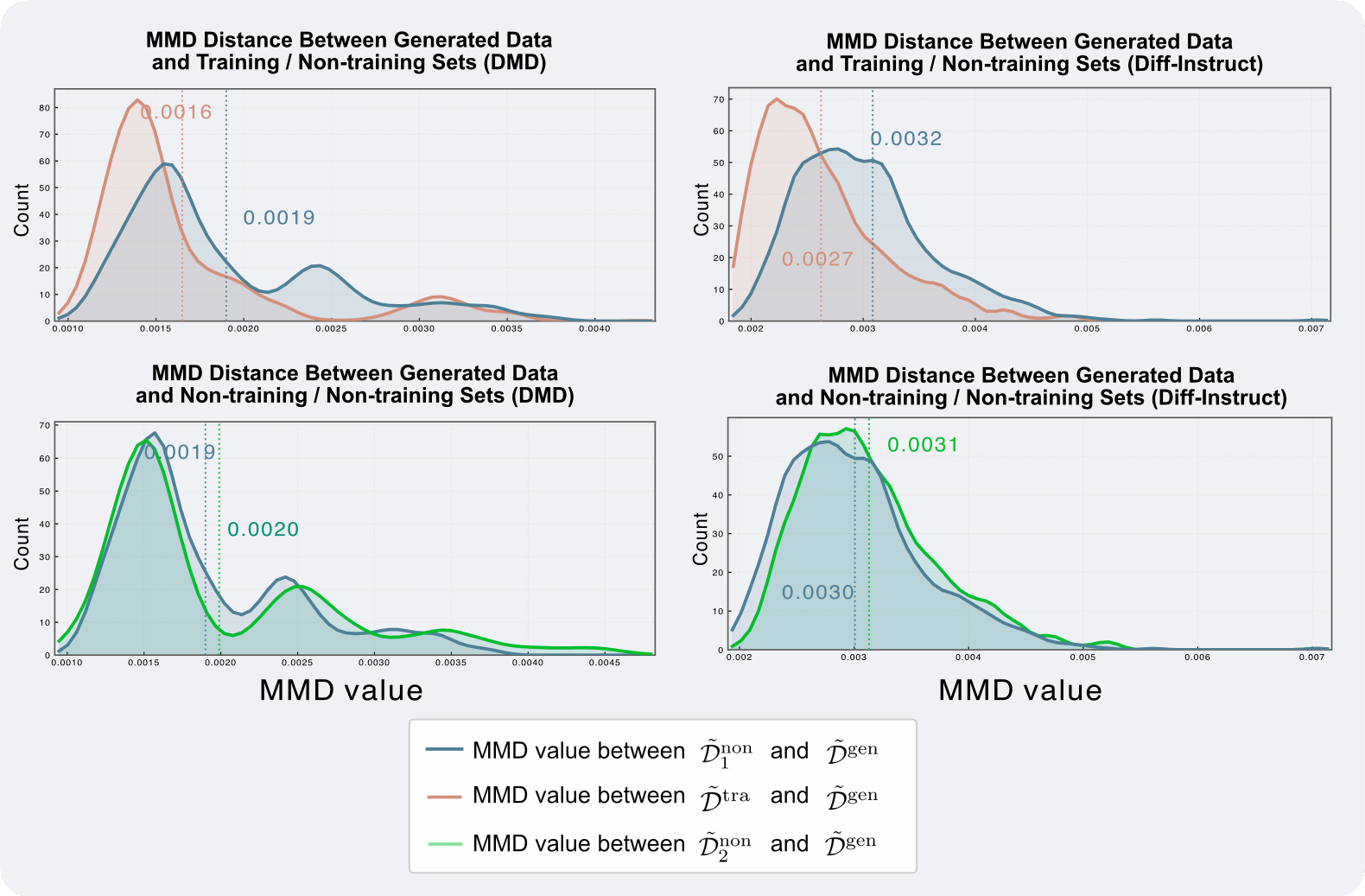}
    \caption{\textbf{The distributional memory chain.} Consistently across different CIFAR-10 student models (e.g., DMD and Diff-Instruct), the MMD between student-generated subset $\mathcal{\tilde{D}}^{\rm gen}$ and the teacher's training subset $\mathcal{\tilde{D}}^{\rm tra}$ concentrates at systematically lower values than the distance to non-training reference subsets $\mathcal{\tilde{D}}^{\rm non}_1$. The lower shift is also reflected by the dashed vertical lines, which mark the mean of each MMD distribution.
    This confirms that a distribution-level trace survives the teacher-student transformation. Conversely, if another non-training subset $\mathcal{\tilde{D}}_2^{\rm non}$ ($\tilde{\gD}^{\rm non}_1 \cap \tilde{\gD}^{\rm non}_2 = \emptyset$) is evaluated as the candidate (as depicted in the bottom ``Non-training / Non-training" plots), its MMD distribution with the generated data ($\mathcal{\tilde{D}}_2^{\rm non}$ and $\mathcal{\tilde{D}}^{\rm gen}$) heavily overlaps with that of the reference non-training set ($\mathcal{\tilde{D}}_1^{\rm non}$ and $\mathcal{\tilde{D}}^{\rm gen}$). The lack of difference between these scenarios further validates that the markedly lower MMD is a signature of actual use of the training data.}
    \label{fig:3}
\end{figure}

The failure is mechanistic, as distillation interposes a transformation layer between the deployed model $G_{\rm S}$ and the original training data $\gD^{\rm tra}$, severing the causal route by which a training image leaves a memorization footprint on the deployed model.
Once $G_{\rm S}$ is trained on teacher-generated samples, the question ``does $G_{\rm S}$ denoise $\vx^* \in \gD^{\rm can}$ unusually well?'' no longer tracks a direct optimization link between $\gD^{\rm can}$ and $G_{\rm S}$, and becomes an operationally ill-posed question in student-only audits, even if the teacher $G_{\rm T}$ was trained on unauthorized data.
In other words, the lack of per-instance memorization on $G_{\rm S}$ results from no instance-level exposure to $\gD^{\rm tra}$; it cannot exonerate $G_{\rm T}$ from unauthorized training with instances within $\gD^{\rm can}$.

\paragraph{A distributional trace survives the teacher-student transformation.}

Although $G_{\rm S}$ does not observe $\gD^{\rm tra}$, it was trained on outputs of $G_{\rm T}$ to approximate the distribution from which $\gD^{\rm tra}$ was drawn~\citep{hinton2015distilling,yin2024one,luo2024diff}.
The student $G_{\rm S}$ may inherit a distributional bias towards the teacher's training data despite having no instance-level exposure to it.
We tested whether such a trace exists by comparing student-generated samples $\gD^{\rm gen}$, the teacher's training set $\gD^{\rm tra}$, and a disjoint reference set $\gD^{\rm non}$.

In each independent trial, we drew random subsets 
$\tilde{\gD}^{\rm gen} \subset \gD^{\rm gen}$, 
$\tilde{\gD}^{\rm tra} \subset \gD^{\rm tra}$, and 
$\tilde{\gD}^{\rm non}_1, \tilde{\gD}^{\rm non}_2 \subset \gD^{\rm non}$, 
where $\tilde{\gD}^{\rm non}_1 \cap \tilde{\gD}^{\rm non}_2 = \emptyset$, 
and measured their pairwise distances using MMD~\citep{gretton2012kernel}.
Across trials, $\mathrm{MMD}(\tilde{\gD}^{\rm gen},\,\tilde{\gD}^{\rm tra})$ 
concentrates at lower values than 
$\mathrm{MMD}(\tilde{\gD}^{\rm gen}, \tilde{\gD}^{\rm non})$ 
in both DMD and Diff-Instruct student models (Fig.~\ref{fig:3}), 
placing the student's outputs consistently closer to the training distribution than to the non-training reference. 
If the candidate set is a non-training set, 
$\mathrm{MMD}(\tilde{\gD}^{\rm gen},\,\tilde{\gD}^{\rm non}_1)$ and 
$\mathrm{MMD}(\tilde{\gD}^{\rm gen}, \tilde{\gD}^{\rm non}_2)$ 
show almost no difference. 
The gap between 
$\mathrm{MMD}(\tilde{\gD}^{\rm gen},\tilde{\gD}^{\rm tra})$ 
and 
$\mathrm{MMD}(\tilde{\gD}^{\rm gen},\tilde{\gD}^{\rm non})$  remains stable across trials, pointing to a systematic bias propagated through the teacher–student pipeline.

This resolves the paradox of upstream training-data influence created by distillation.
While $G_{\rm S}$ does not memorize individual instances of $\gD^{\rm tra}$, its output aligns with the teacher's training data distributionally. 
Therefore, distillation changes the granularity of training-data influence instead of eliminating it. 
Unauthorized data usage leaves a {\em distribution-level memory chain} linking the student's outputs back to $\gD^{\rm tra}$, which becomes more detectable when evidence is aggregated over samples rather than evaluated instance by instance.
This raises an immediate follow-up: can this surviving trace be turned into provenance audits of a deployed student model?

\paragraph{Distributional unauthorized training data detection recovers auditability in distilled models.}

The distributional trace urges a different audit question: not whether $G_{\rm S}$ memorized a particular image, but whether the candidate data is distributionally closer to the student's outputs than a non-training reference is.
We developed {\em distributional unauthorized training data detection}~(D-UTDD) accordingly.
Given the disputed candidate set $\gD^{\rm can}$, a held-out non-training reference set $\gD^{\rm non}$, and a student generation set $\gD^{\rm gen}$ that approximates the training data $\gD^{\rm tra}$, D-UTDD draws random subsets (denoted by $\gB(\cdot)$ for each set) in repeated trials, compares ${\rm MMD}(\gB^{\rm gen}, \gB^{\rm can})$ against ${\rm MMD}(\gB^{\rm gen}, \gB^{\rm non})$, and returns the fraction of of trials in which the candidate (sub)set lies closer to student
generations than the reference does.

We evaluated D-UTDD on CIFAR-10, FFHQ, and AFHQv2, auditing student models distilled via DMD~\citep{yin2024one} and Diff-Instruct~\citep{luo2024diff}.

When $\gD^{\rm can}$ is composed of $\gD^{\rm tra}$, D-UTDD successfully separates $\gD^{\rm can}$ from $\gD^{\rm non}$ with near-perfect performance (with 100\% detection accuracy, ACC and 100\% Area Under the Curve, AUC), while instance-level UTDDs (SecMI~\citep{duan2023diffusion} and ReDiffuse~\citep{li2024towards}) remain close to chance (Fig.~\ref{fig:4}\textbf{a}).
Even under the smallest dataset setting ($|\gD^{\rm non}| = 1200$, $|\gD^{\rm can}| = 600$), where limited sample sizes may weaken the effectiveness of distributional statistics, D-UTDD still maintains robust detection performance (ACC = 83\%).

By aligning the audit mechanism (focusing on distributional statistics) to the form of provenance evidence surviving distillation, D-UTDD recovers the auditability that instance-level UTDDs lose and closes the audit gap created by model distillation.

\paragraph{Quantitative auditing of partial data misuse.}
Real-world copyright disputes rarely involve a model trained exclusively on unauthorized data; more often, a provider can mix a minority of disputed samples into a larger legitimate pool.
We probed this regime by constructing $\gD^{\rm can}$ with 30\%, 50\%, and 100\% training-data fractions.
D‑UTDD maintains high detection rates (e.g., ACC $\approx$ 92\%) even when training data constitute only 30\% of the candidate dataset for both students, whereas instance-level baselines degrade rapidly as the training-data fraction decreases (Fig.~\ref{fig:challenge}).
This robustness is consistent with the distributional observation established in Fig.~\ref{fig:4}\textbf{b}: probing instances collectively amplifies the provenance trace and converts weak per-instance signals into a reliable distribution-level evidence.

Lastly, the D-UTDD audit score shifts monotonically with the training-data proportion in $\gD^{\rm can}$, approaching 1.0 for pure training sets and falling toward 0.5 as prevalence decreases ({Fig.~\ref{fig:4}\textbf{c}).
This suggests that D-UTDD naturally enables another practically-relevant provenance audit scenario: not just \emph{whether} a dataset was used, but \emph{how much} of a suspected dataset is composed of unauthorized data.

\begin{figure}[H]
    \centering
    \includegraphics[width=1\linewidth]{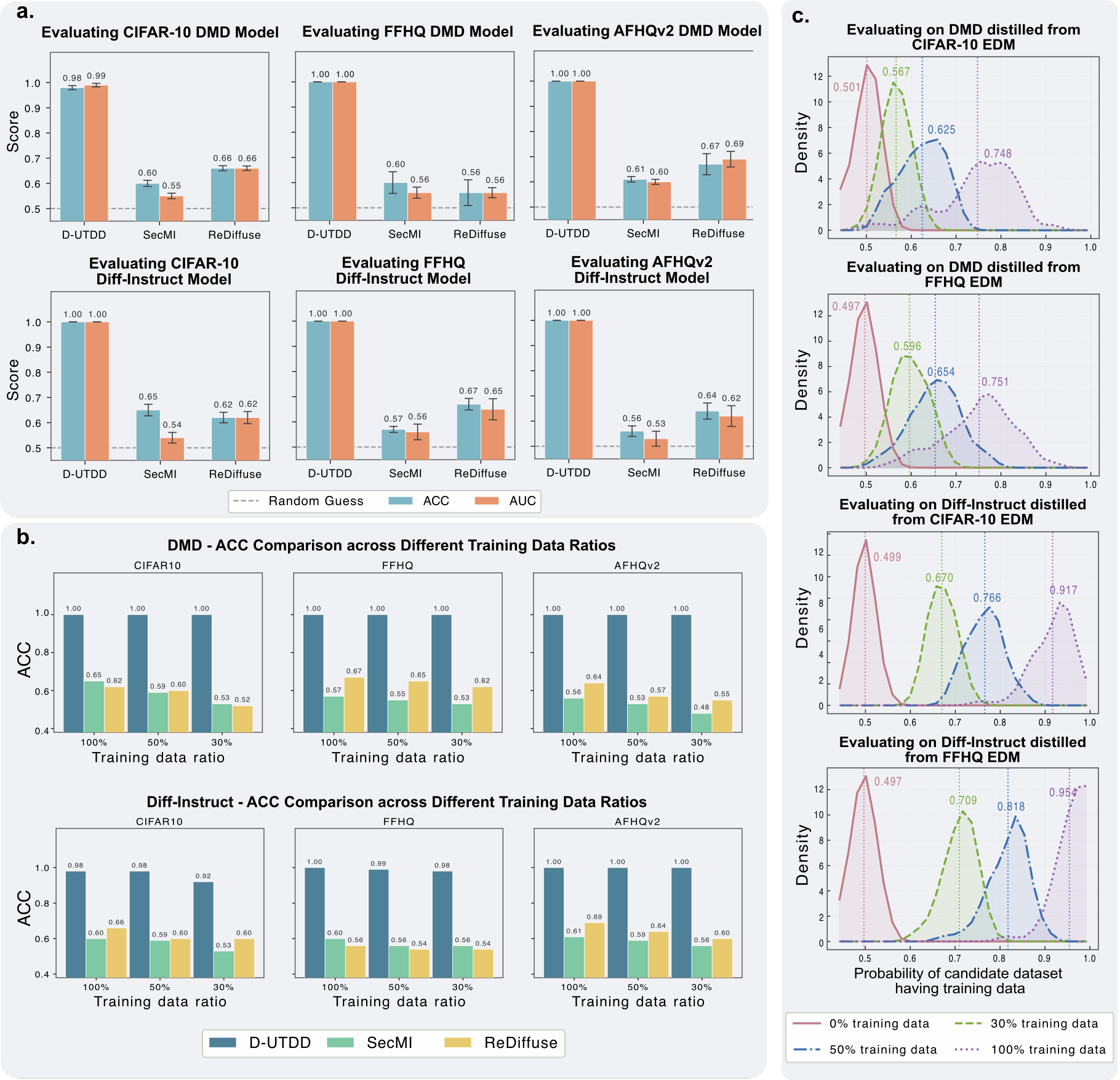}
    \caption{
    \textbf{Recovering auditability in distilled models via D-UTDD.} 
\textbf{a}, Detection on Distilled Students. When evaluating student models distilled via DMD and Diff-Instruct across the CIFAR-10, FFHQ, and AFHQv2 datasets, D-UTDD successfully separates the candidate training set $\mathcal{D}^{\rm can}$ from the non-training reference $\mathcal{D}^{\rm non}$ with near-perfect performance (ACC $\approx$ 100\%). In contrast, state-of-the-art instance-level UTDDs (SecMI and ReDiffuse) fail to find provenance traces, remaining close to random chance.
\textbf{b}, Robustness to Partial Training Data Misuse. In realistic scenarios where the disputed training data constitutes only a minority of the candidate dataset (e.g., training data ratios of 100\%, 50\%, and 30\%), D-UTDD maintains highly robust detection rates. Conversely, the performance of instance-level baselines degrades rapidly as the training-data fraction decreases.
\textbf{c}, Quantitative Provenance Auditing. The D-UTDD audit score shifts monotonically with the proportion of training data in $\mathcal{D}^{\rm can}$, approaching 1.0 for pure training sets and scaling downward as the prevalence decreases. }
    \label{fig:4}
\end{figure}

\section{Discussion}
\label{sec: 6_limitation}

\paragraph{Audit should shift from instance memorization to distributional influence.}

The prevailing training-data provenance auditing has treated instance-level
memorization as both the mechanism and the evidence of unauthorized
use.
Our results expose a structural fragility in this framing. 
Distillation filters out ``fingerprints'' that instance-level UTDDs exploit, while preserving the distributional structure of the original data.
In generative model cascades, the memory chain linking a deployed model to upstream training data is therefore not broken but transformed---from memorization of isolated instances into a statistical prior over the training distribution. 
Audit methodology should evolve to match: from detecting ``reproduction of specific samples'' to detecting ``inheritance of distributional properties''.

\paragraph{Provenance audit in model cascades.}

As generative AI matures into a modular supply chain, where foundation models (teachers) are compressed, distilled, or fine-tuned into specialized endpoints (students), the ``chain of custody'' for training data becomes increasingly opaque. 
Our results indicate that distillation, typically deployed for inference efficiency, incidentally functions as a mechanism for provenance obfuscation: a student model deployed by a downstream provider may ``appear clean'' to conventional inspections, despite being derived from a teacher trained on contested data.

D-UTDD addresses this by treating the student model as a statistical sampler of its ancestor's training distribution, providing a probabilistic basis for establishing {\em upstream influence}, directly relevant to scenarios where unauthorized data have been laundered through synthetic generation.
However, we caution that statistical alignment constitutes evidence of influence, not necessarily legal proof of infringement. 
The distinction between ``learning a style'' (may be legally permissible) and ``appropriating a protected corpus'' remains a complex intersection of statistics and law that our method informs but does not resolve.

\paragraph{Privacy-preserving accountability.}

Paradoxically, the shift to set-based auditing offers a security advantage.
Conventional instance-level UTDDs carry the dual risk of exposing the very private data they aim to protect. 
By attempting to confirm whether a specific record was used in training, these tools can be weaponized to compromise privacy~\citep{carlini2022membership, carlini2023extracting}.
In contrast, distribution-level auditing sidesteps this tension by assessing the collective presence of a dataset without necessarily pinpointing whether any individual element was used for training. 
Although this does not guarantee individual-level privacy, it still holds model developers accountable for the dataset usage as a whole, suggesting that future auditing frameworks can balance rigorous provenance tracking with the minimization of individual privacy risks.

\section{Limitations and Future Outlook}

\paragraph{The Law of Large Numbers constraint.}

The efficacy of distributional audits is bound by the laws of statistics. 
D-UTDD relies on estimating the divergence between high-dimensional distributions (via MMD), and the power of this estimate scales with sample count. 
Detection reliability degrades when the candidate dataset shrinks.

The method is therefore well-suited for auditing systemic dataset usage (e.g., ``Was the portfolio of Artist X used?''), but fails to provide sufficiently strong distributional signals for adjudicating grievances regarding a single image. 
This limitation implies that post-hoc statistical auditing is not a silver bullet. 
For the protection of individual assets, pre-hoc solutions such as robust watermarking~\citep{xu2026antidistillation, zhao2022distillation} likely remain necessary complements.

\paragraph{The reference data assumption.}

Our experimental design assumed access to a non-training reference dataset disjoint from training data.
In wild deployment scenarios where training-set composition is rarely disclosed, constructing a ``clean'' reference is non-trivial.
Thus, developing reference-free or self-calibrated auditing statistics is an
important direction for practical deployment.

\paragraph{Scaling to open-world generators.}
We validated D-UTDD on standard image benchmarks. 
Extending it to text-to-image models trained on billions of image-text pairs introduces additional complexity. 
The relevant distribution may be conditioned on prompts, styles, or semantic categories rather than defined globally. 
Moving from unconditional statistics to prompt- or class-conditioned distributional analysis is a natural next step.

\section{Conclusion}

Distillation is reshaping how generative systems learn, deploy, and forget.
We showed that it erases the instance-level traces current auditors follow yet leaves a durable distributional trace.
By shifting from per-image decision to set-based statistical analysis, D-UTDD recovered the ability to trace data provenance through generative model cascades, which offers a foundation for accountable deployment of distilled generators.
\section{Methods}
\label{sec:Method}
In this section, we detail the setting for all experiments.

\subsection{Threat Model}

\paragraph{Problem setting.}
We consider the problem of auditing whether a deployed image generation service has been trained using unauthorized data under a diffusion model distillation pipeline. 
In this scenario, a generation provider first trains a high-fidelity teacher diffusion model $G_{\rm T}$ on a large dataset $\gD^{\rm tra}$ that may contain unauthorized data. To reduce inference latency, the generation provider distills the trained $G_{\rm T}$ into a one-step student generator $G_{\rm S}$~\citep{yin2024one, luo2024diff}. In practical deployments, $G_{\rm S}$ is the only model released to users as the public-facing generation service, while the $G_{\rm T}$ and its $\gD^{\rm tra}$ remain private and inaccessible. 
Importantly, the $G_{\rm S}$ can be trained using synthetic samples generated by $G_{\rm T}$, without requiring direct access to $\gD^{\rm tra}$.
As a result, the deployed model $G_{\rm S}$ accessible to external parties may not be directly trained on $\gD^{\rm tra}$.

\paragraph{Auditor access and audit objective.}
Under the above setting, since $G_{\rm S}$ is the model released to users, we assume the auditor has white-box access to the deployed student generator $G_{\rm S}$. By querying the model, the auditor can obtain a dataset of generated samples $\gD^{\rm gen} \sim G_{\rm S}$. The auditor does not have access to $G_{\rm T}$, $\gD^{\rm tra}$, or the model training procedure, as the model provider keeps them private and inaccessible.
In addition, the auditor possesses a reference dataset ($\gD^{\rm non}$). We assume that $\gD^{\rm non}$ is drawn from the same underlying distribution as $\gD^{\rm can}$, but is guaranteed not to have been used for model training. $\gD^{\rm non}$ serves as a typical non-training held-out data and provides a baseline for comparison with the candidate dataset $\gD^{\rm can}$ to determine whether it exhibits characteristics of training data. 
Such an assumption is standard in existing UTDD studies~\citep{shokri2017membership, choquette2021label, carlini2022membership}.

In this setting, the auditor is given a candidate dataset $\gD^{\rm can}$ with disputed provenance and aims to determine whether it overlaps with $\gD^{\rm tra}$. Formally, the auditing task is to determine whether $\gD^{\rm can} \cap \gD^{\rm tra} \neq \emptyset$, given only white-box access to $G_{\rm S}$, along with $\gD^{\rm gen}$ and $\gD^{\rm non}$.

\subsection{The Distribution-level Memory Chain}
In this section, we describe the experimental pipeline used to analyze how diffusion model distillation affects instance-level UTDDs and to identify the distributional memory chain.

\paragraph{Teacher diffusion models.}
In our experiments, we require a diffusion model whose training data are known so that the provenance of candidate datasets can be evaluated in a controlled setting. In particular, we need access to two types of data: (1) the training dataset used to train the diffusion model, and (2) a non-training dataset that follows the same underlying distribution as the training data. Ensuring that the non-training data share the same distribution as the training data is critical, as it prevents inherent distributional differences from confounding the evaluation results. 
To satisfy this requirement, we train the diffusion models ourselves and construct the corresponding training and non-training datasets accordingly.

In this work, we adopt the EDM architecture~\citep{karras2022elucidating} as the teacher model. EDM reformulates the diffusion process using a continuous noise parameterization, in which the noise level $\sigma$ serves as the continuous variable that controls the standard deviation of Gaussian noise injected into the data. By parameterizing the diffusion process with a continuous $\sigma$, the denoising function varies smoothly across noise scales, leading to more stable training and improved generation quality~\citep{yin2024one, luo2024diff}.
Moreover, most recent one-step diffusion distillation methods are built upon the EDM architecture~\citep{song2023consistency, yin2024one, luo2024diff}. Therefore, we use EDM as the teacher model to ensure better performance and compatibility with existing distillation pipelines.

We train three EDM teacher models using commonly used diffusion-model training datasets: CIFAR-10, FFHQ, and AFHQv2. CIFAR-10 contains 60,000 color images of size $32\times32$ across ten object categories. FFHQ consists of 70,000 human face images of size $512\times512$, while AFHQv2 includes approximately 15,000 animal face images of size $512\times512$ spanning three species: cats, dogs, and wild animals. Due to the high computational cost of training diffusion models at $512\times512$ resolution, we down-sample the FFHQ and AFHQv2 datasets to $128\times128$ before training.
For each dataset, we randomly divide the data into two equal subsets. One subset is used to train the EDM, while the other serves as non-training data in the subsequent auditing experiments. This setup ensures that both training and non-training data originate from the same underlying distribution.

Using these training splits, we train the EDM teacher models that serve as the source models for subsequent distillation and auditing experiments. 
To ensure strong generative performance, we adopt dataset-specific training configurations for the teacher models. For CIFAR-10, the EDM is trained on a single NVIDIA A100 (80GB) GPU  with a batch size of 128 and a learning rate of $1\times10^{-3}$ for five days. For FFHQ, the EDM is trained on four A100 (80GB) GPUs  with a batch size of 256 and a learning rate of $2\times10^{-4}$ for five days. For  AFHQv2, the EDM is trained on two A100 (80GB) GPUs  with a batch size of 128 and a learning rate of $2\times10^{-4}$ for five days.
The model architectures for different datasets follow the dataset-specific configurations provided in the official EDM repository\footnote{\url{https://github.com/NVlabs/edm}}.

\paragraph{Instance-level UTDD on teacher models.}
After training the EDM teacher models, we evaluate existing UTDDs on them to verify their effectiveness and establish a baseline for comparison with distilled student models. We evaluate three representative UTDDs: SecMI~\citep{duan2023diffusion}, ReDiffuse~\citep{li2024towards}, and GSA~\citep{pang2025white}. These methods attempt to determine whether a candidate sample belongs to the training dataset by exploiting instance-level memorization signals exhibited by diffusion models (Fig.~\ref{fig:challenge}\textbf{b}).

SecMI uses the denoising error during the diffusion process as the detection score. Because diffusion models are trained to predict injected noise, training samples typically produce smaller denoising errors than non-training samples.
In our implementation, we follow the EDM formulation with continuous noise levels. We fix the noise level to $\sigma = 0.95$ and inject this noise into the input image. The EDM model then predicts the noise component, and the detection score is computed as the difference between the model-predicted noise and the injected noise. For the remaining training configurations, we follow the original implementation provided by the authors in the official repository\footnote{\url{https://github.com/jinhaoduan/SecMI}}.

ReDiffuse exploits the reconstruction behavior under repeated perturbations as the detection score. The intuition is that training samples typically produce more accurate reconstructions when the same noisy input is reconstructed multiple times, because the model has already fitted these samples during training.
In our implementation, we inject noise at level $\sigma = 0.95$ into the input image and reconstruct it using the EDM model four times. The reconstructed images are averaged to obtain a mean reconstruction, and the reconstruction error between this averaged image and the original input is used as the detection score. For the remaining training configurations, we follow the original implementation provided by the authors in the official repository\footnote{\url{https://github.com/lijingwei0502/diffusion_mia?tab=readme-ov-file}}.

GSA uses the size of the parameter updates during retraining as the detection score. The intuition is that training samples typically induce smaller parameter updates when the model is fine-tuned on them, as the model has already fitted these samples during training. 
In our implementation, we fine-tune the trained EDM model on candidate samples and measure the size of parameter updates during retraining. Training samples tend to produce smaller gradient updates compared with non-training samples. For the training setup, we use hyperparameter settings consistent with those reported in the original paper. The implementation details are available in the official repository\footnote{\url{https://github.com/py85252876/GSA}}.

In addition to the configurations described above, these instance-level UTDDs require a decision threshold to distinguish training samples from non-training samples based on their detection scores. Specifically, each method assigns a score to a candidate sample that reflects the likelihood of it belonging to the training dataset. A threshold is then used to classify samples as training or non-training according to this score.
We determine the threshold using a validation set consisting of both training and non-training samples. The threshold is selected to best separate the detection scores of training and non-training samples.

\paragraph{Distillation pipeline.}
We further distill student generators from the trained EDM teacher models and evaluate existing UTDDs on these models to investigate whether instance-level memorization signals remain detectable after diffusion model distillation. We adopt two representative one-step diffusion distillation methods, DMD~\citep{yin2024one} and Diff-Instruct~\citep{luo2024diff}, to distill student models from the trained EDMs.

These distillation methods rely solely on synthetic samples generated by the teacher model, allowing the student model to be trained without direct access to the $\gD^{\rm tra}$.
To distill student models, we first construct synthetic training datasets using the trained EDM teacher models. Specifically, we randomly sample Gaussian noise and use the trained EDM teacher models to generate the corresponding images, forming noise–image training pairs for supervising the student generators.
In total, each teacher model produces 100,000 synthetic samples. During distillation, the student models learn to directly map the input noise to the corresponding generated image in a single forward pass. Different distillation approaches adopt different loss functions to supervise this learning process, but they all rely on the same synthetic noise–image pairs as the training data.

Using these synthetic data, we then distill the student generators. 
The training hyperparameters are configured separately for each distillation method and dataset.
For the DMD approach, training is conducted on a single NVIDIA A100 GPU (80GB) for four days. The batch size is set to 128 for CIFAR-10 and 64 for both FFHQ and AFHQv2, with a learning rate of $5\times10^{-5}$ across all datasets.
For the Diff-Instruct approach, training is conducted on a single NVIDIA A100 (80GB) GPU, with training durations varying across datasets: two days for CIFAR-10 and FFHQ, and two days for AFHQv2. The batch size follows the same configuration as DMD (128 for CIFAR-10 and 64 for FFHQ and AFHQv2), and the learning rate is set to $1\times10^{-5}$ for CIFAR-10 and $1\times10^{-4}$ for FFHQ and AFHQv2.
Following the training strategies proposed in DMD and Diff-Instruct, the student models adopt the same architectures as their corresponding teacher models~\citep{yin2024one, luo2024diff}.

\paragraph{Instance-level UTDD on student models.}
To examine whether existing instance-level UTDDs remain effective after model distillation, we apply the same detectors—SecMI, ReDiffuse, and GSA—to the distilled student generators. The evaluation protocol largely follows the setup used for the trained EDM teacher model.
For SecMI and ReDiffuse, which rely on injecting noise into input images, we adjust the noise level when evaluating the student models. We adopt a noise level of $\sigma = 0.27$ for the student-model evaluation. Through preliminary experiments, we find that noise levels around $\sigma = 0.27$ produce relatively stronger detection signals on the student generators. Other implementation details remain consistent with the settings described in the previous section. Despite extensive attempts to tune the evaluation settings, these three UTDDs still fail to reliably distinguish training and non-training samples on the distilled student models.

\paragraph{Distributional trace.}
Although model distillation largely obscures the instance-level memorization signals exploited by existing UTDDs, our experiments reveal that distribution-level traces of the teacher’s training data can still survive after distillation (Fig.~\ref{fig:3}). Specifically, we observe that the data generated ($\gD^{\rm gen}$) by $G_{\rm S}$ may inherit a distributional bias toward the teacher’s training dataset, even though the student model has never directly accessed those training samples.

This observation is obtained through the following experiment. We use $\gD^{\rm gen}$, as an anchor distribution. We then compare the distributional distances between $\gD^{\rm gen}$ and two reference datasets: $\gD^{\rm tra}$ and a $\gD^{\rm non}$. If the student generator indeed inherits distributional traces from the teacher’s training data, we expect $\gD^{\rm gen}$ to be closer to $\gD^{\rm tra}$ than to $\gD^{\rm non}$.

To quantify these distributional distances, we adopt \textit{Maximum Mean Discrepancy} (MMD), a widely used statistical measure for comparing probability distributions. 
Formally, MMD, proposed by~\citet{gretton2012kernel}, measures the distance between two Borel probability measures $\sP$ and $\sQ$ defined on a separable metric space $\gX \subseteq \mathbb{R}^d$.
Consider independent random variables $X, X^\prime \sim \sP$ and $Y, Y^\prime \sim \sQ$.
The squared MMD between $\sP$ and $\sQ$ in a Reproducing Kernel Hilbert Space $\sH_k$, induced by a kernel function $k: \gX \times \gX \to \sR$, is defined as:
\begin{equation*}
    \operatorname{MMD}^2(\mathbb{P}, \mathbb{Q}; k) = \mathbb{E}[k(X, X')] + \mathbb{E}[k(Y, Y')] - 2\mathbb{E}[k(X, Y)].
\end{equation*}
If $k$ is a characteristic kernel (e.g., Gaussian), then $\operatorname{MMD}^2(\mathbb{P}, \mathbb{Q}; k) = 0$ if and only if $\sP = \sQ$.

In practice, the true distributions $\sP$ and $\sQ$ are often unknown, and we rely on finite samples drawn from them.
Given i.i.d. samples $\gS_{X} = \{ x_i \}_{i=1}^{n}$ from $\sP$ and $\gS_{Y} = \{ y_j \}_{i=1}^{m}$ from $\sQ$, an unbiased U-statistic estimator for $\operatorname{MMD}^2$ is
\begin{equation}
\label{eq:mmd_unbiased_est}
\begin{split}
\widehat{\operatorname{MMD}}_u^2(\gS_X, \gS_Y; k) 
&= \frac{1}{n(n-1)} \sum_{i \neq l}^{n} k(x_i, x_l)
 + \frac{1}{m(m-1)} \sum_{j \neq p}^{m} k(y_j, y_p) \\
&\quad - \frac{2}{nm} \sum_{i=1}^{n} \sum_{j=1}^{m} k(x_i, y_j).
\end{split}
\end{equation}

In our experiments, the kernel $k$ is instantiated as a Gaussian kernel,
\begin{equation}
k(x,y) = \exp\left(-\frac{\|x-y\|^2}{2\alpha^2}\right),
\end{equation}
where $\alpha$ is the kernel bandwidth. $\alpha$ is an important hyperparameter, as it controls the sensitivity of the kernel to differences between samples: smaller bandwidths focus on local discrepancies, while larger bandwidths capture more global distributional differences.

Instead of fixing this parameter manually, we determine $\alpha$ adaptively based on the empirical data distribution. Specifically, for each pair of datasets being compared, the bandwidth is computed as the average pairwise Euclidean distance within the two feature sets:
\begin{equation}
\alpha
=
\frac{1}{2}
\left(
\frac{\sum_{i \ne j} \|x_i - x_j\|}{n(n-1)}
+
\frac{\sum_{i \ne j} \|y_i - y_j\|}{m(m-1)}
\right),
\end{equation}
where the two terms correspond to the mean intra-set pairwise distances of the two feature sets.

\paragraph{Distributional trace experiment configuration.}
Building on the MMD setup described above, we now specify the experimental configuration used to quantify the distributional trace that survives the teacher–student transformation. First, we construct sampled subsets of size 400 from $\gD^{\rm gen}$, $\gD^{\rm tra}$ and $\gD^{\rm non}_1$, denoted by $\tilde{\gD}^{\rm gen}$, $\tilde{\gD}^{\rm tra}$, and $\tilde{\gD}^{\rm non}_1$, respectively.
To provide a clean control comparison, we randomly sample another non-training subset, denoted by $\tilde{\gD}^{\rm non}_2$, where $\tilde{\gD}^{\rm non}_2 \cap \tilde{\gD}^{\rm non}_1 = \emptyset$. 
Then, we compute the distance gap between $\rm{MMD}(\tilde{\gD}^{\rm gen}, \tilde{\gD}^{\rm tra})$ and $\rm{MMD}(\tilde{\gD}^{\rm gen}, \tilde{\gD}^{\rm non}_1)$, to examine whether the student-generated data remain systematically closer to the teacher’s training distribution than to non-training data. 
As a clean control, we further compute the distance gap between $\rm{MMD}(\tilde{\gD}^{\rm gen}, \tilde{\gD}^{\rm non}_1)$ and $\rm{MMD}(\tilde{\gD}^{\rm gen}, \tilde{\gD}^{\rm non}_2)$. Since both $\tilde{\gD}^{\rm non}_1$ and $\tilde{\gD}^{\rm non}_2$ are drawn from the same non-training distribution, these two distances are expected to show little difference.
Finally, this sampling-and-comparison procedure is repeated over 5000 independent trials, yielding empirical distributions of MMD values.


\subsection{D-UTDD Framework}
Motivated by the distributional memory chain identified in the previous section, we develop D-UTDD, a distribution-level auditing framework designed to detect unauthorized training data in distilled diffusion models. 

\paragraph{Deep-Kernel MMD.}
In the previous section, we showed how the distributional trace can be identified. To quantify this trace, we adopt MMD as the core statistical distance. However, directly applying standard MMD can be suboptimal for high-dimensional data~\citep{liu2020learning} (e.g., image embeddings with thousands of dimensions), where fixed kernels may fail to capture task-relevant semantic differences between distributions. 

To address this, \citet{liu2020learning} propose to learn a task-relevant representation $\theta_{\omega}: \gX \to \gZ$ using a neural network parameterized by $w$.
The MMD can then be computed in this learned feature space $\gZ$.
As such, the goal of deep-kernel MMD is to find a representation $\theta_{\omega}$ that maximizes the MMD, thereby increasing the test power to detect differences between $\sP$ and $\sQ$.
Following~\citep{liu2020learning}, let $\gS_{X} = \{ x_i \}_{i=1}^{n}$ and $\gS_{Y} = \{ y_j \}_{i=1}^{n}$ be samples from $\sP$ and $\sQ$ (assuming equal sample sizes $n$ for simplicity).
The empirical estimate for deep-kernel MMD, using a U-statistic, is
\begin{equation}
    \widehat{\operatorname{MMD}}^2_u(\gS_{X}, \gS_{Y}; k_{\omega}) := \frac{1}{n(n-1)} \sum_{i \neq j} H_{ij},
\end{equation}
where $H_{ij}$ is the kernel of the U-statistic, defined as
\begin{equation}
    H_{ij} := k_{\omega}(x_i, x_j) + k_{\omega}(y_i, y_j) - k_{\omega}(x_i, y_j) - k_{\omega}(y_i, x_j).
\end{equation}
Note that the kernel $k_{\omega}(\cdot,\cdot)$ itself is a composite function incorporating the learned features:
\begin{equation}
k_{\omega}(a, b) = \left [ \left ( 1-\epsilon \right ) \, k_{\rm base} \left( \theta_{\omega}(a), \theta_{\omega}(b) \right) + \epsilon \right] \cdot q_{\rm base} \left (a, b \right).
\end{equation}
Here, $w$ is the feature extractor network (e.g., a multi-layer perceptron). 
$k_{\rm base}(\cdot, \cdot)$ is a base characteristic kernel (e.g., Gaussian) applied to the learned features $\theta_{\omega}(a)$ and $\theta_{\omega}(b)$.
In addition, $q_{\rm base}(\cdot,\cdot)$ is typically another characteristic kernel on the original inputs, acting as a sample-pair weighting function that adjusts the influence of each pair in the kernel computation based on their importance or relevance.
The small constant $\epsilon \in [0, 1]$ helps to ensure $k_{w}$ remains characteristic \citep{liu2020learning}.

When optimizing the parameters $w$ of the feature network $\theta_{\omega}$ to maximize the MMD estimate, it is often normalized by an estimate of its standard deviation to improve numerical stability and test power.
The objective function is thus:

\begin{equation}
    \max_{\omega} \gL({\omega}) = \frac{ \widehat{\operatorname{MMD}}^2_u(\gS_{X}, \gS_{Y}; k_{\omega}) }{ \sigma \left( \widehat{\operatorname{MMD}}^2_u(\gS_{X}, \gS_{Y}; k_{\omega}) \right) },
\end{equation}

where $\sigma(\cdot)$ denotes the standard deviation of the MMD estimator.
Since the true variance $\sigma^2$ is generally unknown, we estimate it using a regularized estimator $\hat{\sigma}_\lambda^2$, given by:

\begin{equation}
    \hat{\sigma}^2_{\lambda} = \frac{4}{n^3} \sum_{i=1}^{n} \left( \sum_{j=1}^{n} H_{ij} \right)^2 
    - \frac{4}{n^4} \left( \sum_{i=1}^{n} \sum_{j=1}^{n} H_{ij} \right)^2 + \lambda,
\end{equation}
where $\lambda$ is a constant to avoid division by zero.
The optimization of $w$ is typically performed using stochastic gradient ascent on $\gL(w)$.

Within the D-UTDD, deep-kernel MMD is used to quantify the distributional discrepancy between datasets and the student-generated distribution.

\begin{figure*}[t]
    \centering
    \includegraphics[width=\linewidth]{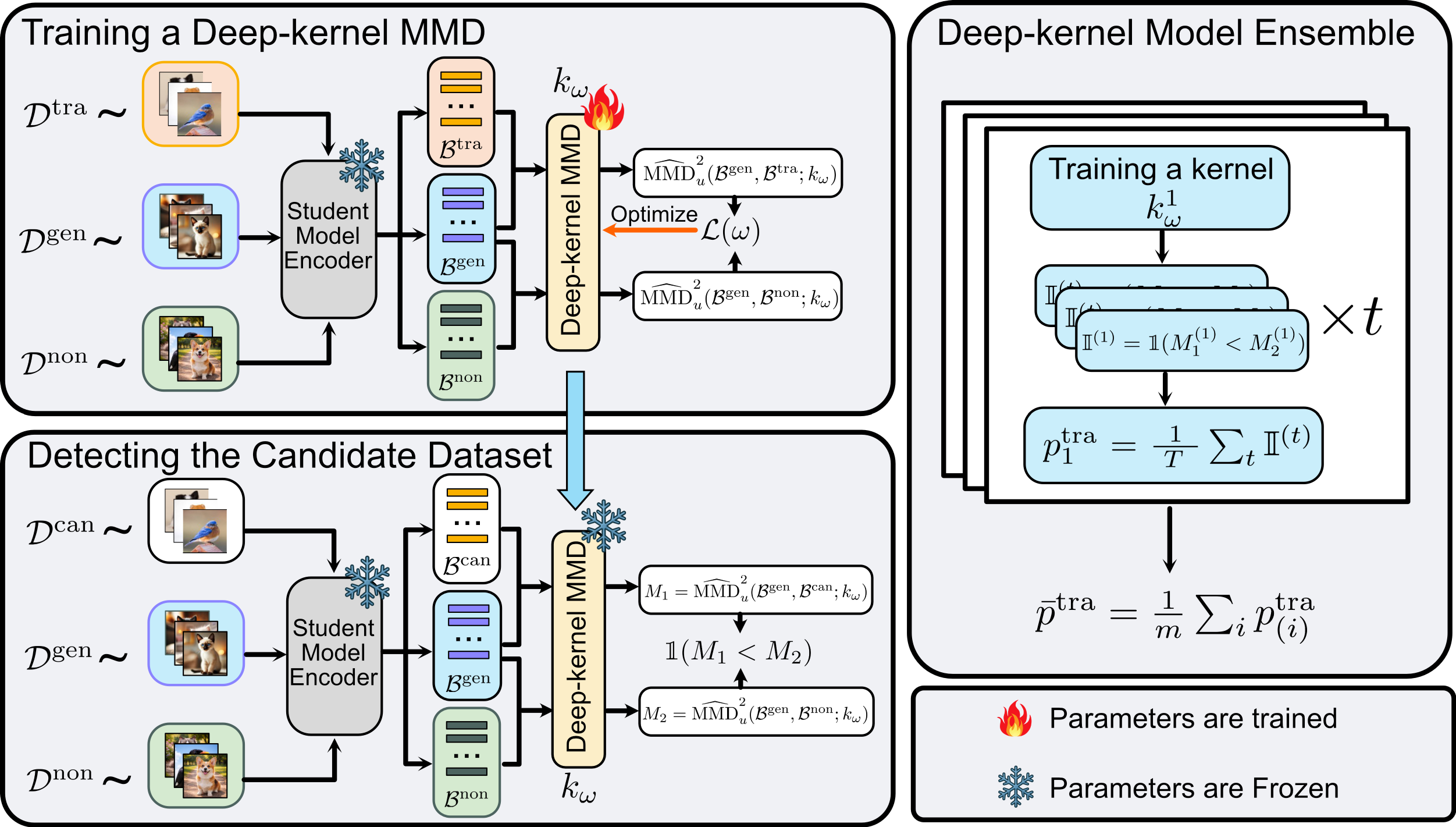} 
    \caption{
    Overview of our D-UTDD pipeline, consisting of (1) \textit{Training a Deep-kernel MMD} stage (top left) and (2) \textit{Detecting the Candidate Dataset} stage (bottom left).
    We also propose a Deep-kernel Model Ensemble strategy to improve detection robustness (right).
    }
    \label{fig:5}
\end{figure*}

\paragraph{Two-stage auditing pipeline.}
Building on the deep-kernel MMD introduced above, D-UTDD performs auditing through a two-stage pipeline consisting of (1) a \emph{deep-kernel MMD training stage} and (2) a \emph{candidate dataset detection stage}.
In the first stage, a deep-kernel is learned to enlarge the gap between the distributional distance from training data to the student-generated distribution and that from non-training data to the student-generated distribution.
In the second stage, the learned kernel is used to evaluate candidate datasets by comparing their distributional distance to the student-generated distribution against that of a reference non-training dataset (Fig.~\ref{fig:5}).

Before entering the two stages, all datasets are first mapped into a shared representation space using the student model's encoder. Specifically, each sample is passed through the student model encoder to obtain a feature embedding, and all subsequent distribution comparisons are performed in this feature space rather than the raw input space.

During the \emph{deep-kernel MMD training} stage, we optimize a data-adaptive kernel $k_{\omega}$ used in deep-kernel MMD, which is parameterized by deep neural networks $\omega$~\citep{liu2020learning}, which are trained on ${\gD}^{\rm tra}$ and $D^{\rm non}$ to maximize the separation between the two datasets in the feature space.
For ${\gD}^{\rm tra}$, $\gD^{\rm non}$ and $\gD^{\rm gen}$, we perform mini-batch training and randomly sample subsets from each dataset, e.g., $\gB^{\rm gen} = \{ \vx_b^{*} \overset{{\rm i.i.d}}{\sim} \gD^{\rm gen} \}_{b=1}^{B}$, with respect to the optimization objective $\gL(\omega)$ defined as
    \begin{equation*}
    \gL(\omega) = \underbrace{\left[ \widehat{\rm MMD}^2_u(\gB^{\rm gen}, {\gB}^{\rm tra}; k_\omega) \right]}_{\text{training data discrepancy}} - 
        \underbrace{ \left[ \widehat{\rm MMD}^2_u (\gB^{\rm gen}, \gB^{\rm non}; k_\omega) \right]}_{\text{non-training data discrepancy}}.
    \end{equation*}
Doing so amplifies the MMD values between non-training data and the student-generated distribution while minimizing them for training-like distributions.

To detect the candidate dataset, in the \emph{detecting the candidate dataset} stage, we aim to determine whether $\gD^{\rm can} \cap \gD^{\rm tra} = \varnothing$, by computing two MMD statistics using the trained kernel $k_{\omega}$: $M_1^{(t)} \triangleq \widehat{\rm MMD}^2_u(\gB^{\rm gen}, \gB^{\rm can}; k_\omega)$ and $M_2^{(t)} \triangleq \widehat{\rm MMD}^2_u(\gB^{\rm gen}, \gB^{\rm non}; k_\omega)$ over $T$ Bernoulli trials.
The prediction result for each trial is indicated via $\sI^{(t)} = \mathbbm{1}(M_1^{(t)} < M_2^{(t)})$, and the aggregate training data probability is estimated by $p^{\rm tra} = \frac{1}{T} \sum_t \sI^{(t)}$.

\paragraph{Ensembling multiple kernels}
While the two-stage auditing pipeline produces a training-data probability for a candidate dataset, the estimate may still suffer from variance due to finite-sample MMD estimation~\citep{cherief2022finite}. To improve robustness, D-UTDD further adopts a kernel ensemble strategy that aggregates the results of multiple independent auditing runs.
Specifically, we train $m$ independent kernels $\{ k_{\omega}^{(i)} \}_{i=1}^{m}$, and apply the two-stage auditing pipeline separately with each kernel. For each kernel, we obtain a training-data probability $p^{\rm tra}_{(i)}$, from the candidate dataset detection stage. The final audit score is computed as the ensemble mean $\bar{p}^{\rm tra} = \frac{1}{m} \sum_i p^{\rm tra}_{(i)}$. A candidate dataset is declared as containing training data if $\bar{p}^{\rm tra} > \tau$.


\paragraph{Practical constraint: synthetic surrogate for inaccessible $\gD^{\rm tra}$.}
In the auditing setting considered in this work, the auditor does not have access to $G_{\rm T}$, and $\gD^{\rm tra}$. In the \textit{deep-kernel MMD training stage}, optimizing the kernel requires training data to contrast against non-training data. Directly assuming access to $\gD^{\rm tra}$ would therefore contradict the auditing setting considered in this work. 
Since the student model is trained to mimic the teacher model’s output distribution and therefore inherits a distributional bias toward the teacher’s training data~\citep{karras2022elucidating}, its generated samples can serve as training-like signals. To address the practical constraint, we use synthetic samples generated by the student model as a surrogate for the training data required for deep-kernel optimization.
Throughout all subsequent D-UTDD experiments, this synthetic surrogate is consistently used in place of the genuine training dataset to ensure that the evaluation remains consistent with the assumed auditing setting.

\subsection{D-UTDD Experimental Protocols}
Building on the D-UTDD pipeline described above, we now introduce the experimental protocols used to implement the method and evaluate its effectiveness.

\paragraph{Evaluation metrics.}
To evaluate the effectiveness of the proposed auditing framework, we adopt standard evaluation metrics commonly used in prior work on UTDD~\citep{carlini2022membership}.
Specifically, we report two metrics: Accuracy (ACC) and the area under the receiver operating characteristic curve (AUC). 
ACC measures the proportion of correctly classified samples in both the training and non-training groups under a fixed decision threshold. Formally,
\begin{equation}
\mathrm{ACC} = \frac{TP + TN}{TP + TN + FP + FN},
\end{equation}
where $TP$ and $TN$ denote the numbers of correctly identified training and non-training samples, respectively, while $FP$ and $FN$ denote the corresponding misclassification counts.

AUC evaluates the ranking consistency between training and non-training scores. It can be interpreted as the probability that a randomly selected training sample receives a higher detection score than a randomly selected non-training sample.

\paragraph{Datasets and data partitioning.}
As described in the D-UTDD framework, the auditing procedure requires three types of datasets: (1) training data $\gD^{\rm tra}$ and (2) non-training reference data $\gD^{\rm non}$ used for deep-kernel optimization and candidate data detection, (3) student-generated samples $\gD^{\rm gen}$ used as the anchor distribution. Following the practical auditing setting introduced earlier, genuine training data are not accessible during the auditing procedure and are replaced by synthetic samples generated $\hat{\gD}^{\rm tra}$ by the student model.

In the \textit{deep-kernel MMD training} stage, we construct synthetic training datasets $\hat{\gD}^{\rm tra}$ using samples generated by the student model. We also randomly sample a subset of non-training data to construct non-training reference data $\gD^{\rm non}$.
Specifically, for CIFAR-10 and FFHQ, we construct 5000 synthetic training samples and 5000 non-training samples. For AFHQv2, which has a smaller dataset size, we construct 1500 synthetic training samples and 1,500 non-training samples. For evaluation, we further construct a student-generated dataset $\gD^{\rm gen}$ consisting of 5000 images sampled from the student generator. 

In the \textit{detecting the candidate dataset} stage, we construct candidate datasets $\gD^{\rm can}$ to simulate different provenance scenarios. Each candidate dataset contains 5000 samples (or 1000 samples for AFHQv2 due to its smaller dataset size) obtained through sampling without replacement. Depending on the evaluation setting, the samples are drawn either from the genuine training dataset or from the remaining non-training dataset.
In total, we construct 100 candidate datasets. Among them, 50 datasets consist entirely of genuine training data to simulate the scenario where the candidate dataset contains training data, while the remaining 50 datasets consist entirely of non-training data to represent the scenario where no training data are present.

In addition, we also evaluate mixed-candidate scenarios where samples are drawn from the genuine training data and non-training data according to predefined mixing ratios. For each mixing ratio (30\% and 50\%), 50 candidate datasets are constructed to evaluate the robustness of the auditing method under partial training-data usage.

\paragraph{Baseline UTDDs.}
To compare D-UTDD with existing UTDDs, we evaluate several representative instance-level baselines under the same candidate-dataset setting described above. Since these existing UTDDs are designed for instance-level detection, we slightly adapt their evaluation protocol to enable dataset-level auditing.
Specifically, we first sample both training and non-training data to determine an appropriate decision threshold for each baseline method. The threshold is selected to best separate the detection scores of training and non-training samples. During evaluation, the baseline detector is applied to all samples in $\gD^{\rm can}$. For each sample, the detection score is computed using the same procedure and hyperparameter settings as in the previous instance-level UTDD evaluation of the student models.
If more than 50\% (or 30\% in the mixed candidate dataset setting, where the training data proportion is 30\%) of the sample are classified as training data at the selected threshold, the candidate dataset is considered to contain training data.
This adaptation allows instance-level UTDDs to be evaluated under the same dataset-level auditing scenario as D-UTDD, enabling a fair comparison of their effectiveness in detecting the presence of training data within candidate datasets.

\paragraph{Validation of deep-Kernel MMD configuration.}
To obtain stable and effective configurations for the deep-kernel MMD, we validate the kernel network and its hyperparameters for each dataset and distillation model.
In deep-kernel MMD, the kernel function operates on a learned feature representation produced by a neural feature extractor. 
In our implementation, this extractor is a fully connected neural network that maps high-dimensional input embeddings to a lower-dimensional representation before computing the kernel discrepancy. The architecture is parameterized by two variables: $H$, the dimensionality of all hidden layers, and $x_{\rm{out}}$, the dimensionality of the output embedding. The resulting features are then used with a Gaussian kernel, whose bandwidth serves as an additional hyperparameter that controls the scale of similarity. This design follows the standard deep-kernel MMD architecture introduced by \citet{liu2020learning}.

To determine suitable hyperparameters, we perform a validation procedure for each dataset–model pair. Specifically, we randomly sample 2,000 synthetic training samples and 2,000 non-training samples to construct validation pairs. For AFHQv2, due to its smaller dataset size, we instead sample 800 synthetic training samples and 800 non-training samples. We perform a greedy search over five hyperparameters: the kernel bandwidth, the learning rate of the feature extractor, the number of training epochs, and the architectural parameters $H$ and $x_{\rm{out}}$. At each step, we select the configuration that maximizes the separability between training-like and non-training data before proceeding to the next parameter dimension.

For the Diff-Instruct model, the search yields the following configurations.
On CIFAR-10, the optimal setting uses a kernel bandwidth of $0.1$, $400$ training epochs, and a learning rate of $1\times10^{-6}$, with a feature extractor consisting of $H=450$ hidden units and an output dimension of $x_{\text{out}}=35$.
On FFHQ, the optimal configuration adopts a bandwidth of $0.4$, $300$ epochs, the same learning rate of $1\times10^{-6}$, and a feature extractor with $H=450$ and $x_{\text{out}}=50$.
On AFHQv2, the best configuration uses a bandwidth of $0.1$, $400$ epochs, a learning rate of $1\times10^{-6}$, and $H=450$ with $x_{\text{out}}=35$.

For the DMD model, the optimal configuration differs slightly.
On CIFAR-10, the best setting uses a smaller bandwidth of $0.0025$, $300$ epochs, and a learning rate of $1\times10^{-7}$, together with $H=250$ and $x_{\text{out}}=20$.
For FFHQ and AFHQv2, the optimal configurations are consistent with those used for Diff-Instruct, adopting bandwidths of $0.4$ and $0.1$ respectively, $300$ or $400$ training epochs depending on the dataset, a learning rate of $1\times10^{-6}$, and a feature extractor with $H=450$ and $x_{\text{out}}=50$ or $35$.

\paragraph{D-UTDD configuration.}
Building on the validated deep-kernel MMD configuration described above, we now specify the experimental setup used to implement the D-UTDD auditing pipeline. The configuration follows the two-stage design introduced before, consisting of a deep-kernel training stage and a candidate dataset detection stage. 

A key parameter in this setup is the batch size of the sampled subsets, which is used both in deep-kernel training and in the subsequent distributional comparisons.
During the deep-kernel training stage, we randomly sample 400 synthetic training samples, 400 non-training samples, and 400 generated samples, forming the batches $\gB^{\rm tra}$, $\gB^{\rm non}$, and $\gB^{\rm gen}$, respectively. These samples are fed into the target model, and feature representations are extracted from an intermediate layer of the model. 
In the candidate dataset detection stage, we repeatedly draw random subsets of 400 samples from the candidate dataset $\gD^{\rm can}$ to construct batches $\gB^{\rm can}$. Their corresponding feature representations are extracted in the same manner. By comparing the feature distribution of $\gB^{\rm can}$ against those of $\gB^{\rm non}$ and $B^{\rm gen}$ using the trained deep-kernel MMD.

Following the ensembling strategy, we further aggregate multiple independent auditing runs to improve robustness. Specifically, for each trained kernel, we repeat the candidate dataset detection procedure over $t$ Bernoulli trials to estimate a training-data probability $p^{\rm tra}_{(i)}$. We then train $m$ independent kernels and apply the two-stage auditing pipeline separately with each kernel. The final audit score is computed as the ensemble mean $\bar{p}^{\rm tra}$ over the resulting $m$ training-data probabilities. In our experiments, we set $t = 100$ Bernoulli trials for each kernel and $m = 10$ kernels for ensembling. 

A decision threshold is further required because D-UTDD produces an audit score, namely the ensemble mean training-data probability $\bar p^{\rm tra}$. In principle, a score around 0.5 corresponds to an ambiguous case where the candidate dataset is not consistently closer to the student-generated distribution than the non-training reference. 
To avoid making overly sensitive decisions near this boundary, we adopt a fixed threshold slightly above the chance level. In all D-UTDD experiments, we set the threshold to 0.54.





\newpage
\subsection*{Data Availability}
Public datasets used in this study, including CIFAR-10, FFHQ, and AFHQv2, are available from their original sources. The dataset versions, preprocessing procedures, split definitions, and candidate-set construction rules used in this work are provided in the accompanying repository. Source data underlying the main figures and tables are also provided. Where raw data cannot be redistributed due to licensing or copyright restrictions, we provide the metadata, split files, and processed outputs necessary to reproduce the reported analyses.


\subsection*{Author Contributions Statement}
Muxing Li, Zesheng Ye, Feng Liu, and Sharon Li conceived the idea. Muxing Li and Zesheng Ye conducted the experiments and implemented the code. Muxing Li, Zesheng Ye, and Feng Liu prepared the first draft of the manuscript. All authors contributed to multiple rounds of manuscript revision and approved the final version.

\subsection*{Funding Statement}
Muxing Li, Guangquan Zhang, and Feng Liu were supported by the Australian Research Council with grant DP230101540. Feng Liu was also supported by the Australian Research Council with grant DE240101089.
\newpage
\bibliography{main}
\clearpage

\end{document}